\UseRawInputEncoding
\documentclass[a4paper,fleqn]{cas-dc}

\usepackage[numbers]{natbib}

\usepackage{framed} 
\usepackage{amsmath} 
\usepackage{amsfonts}
\usepackage{graphicx} 
\usepackage{algorithm}
\usepackage{algorithmic}
\usepackage{hyperref} 
\usepackage{orcidlink}
\usepackage[utf8]{inputenc}

\begin{document}

\title[mode = title]{The Power of Certainty: How Confident Models Lead to Better Segmentation}

\author[inst1]{Tugberk Erol\orcidlink{0000-0002-0831-7307}}
\ead{tugberk.erol@gazi.edu.tr}

\author[inst2]{Tuba Caglikantar\orcidlink{0000-0001-5590-5307}}
\ead{tubacaglikantar@gmail.com}

\author[inst3]{Duygu Sarikaya\orcidlink{0000-0002-2083-4999}}
\ead{d.sarikaya@leeds.ac.uk}

\affiliation[inst1]{organization={Department of Computer Engineering},
            addressline={Graduate School of Natural and Applied Sciences}, 
            city={Gazi University},
            country={Turkey}}

\affiliation[inst2]{organization={Department of Software Engineering},
            addressline={Faculty of Engineering and Natural Sciences}, 
            city={Ankara Yildirim Beyazit University},
            country={Turkey}}

\affiliation[inst3]{organization={School of Computer Science},
            city={University of Leeds},
            country={United Kingdom}}

\begin{keywords}
Self Distillation \sep
Confidence-Based Distillation \sep
Regularization \sep
Polyp segmentation \sep
Medical image segmentation \sep
Convolutional networks \sep
\end{keywords}

\begin{abstract}
 Deep learning models have been proposed for automatic polyp detection and precise segmentation of polyps during colonoscopy procedures. Although these state-of-the-art models achieve high performance, they often require a large number of parameters. Their complexity can make them prone to overfitting, particularly when trained on biased datasets, and can result in poor generalization across diverse datasets. Knowledge distillation and self-distillation are proposed as promising strategies to mitigate the limitations of large, over-parameterized models. These approaches, however, are resource-intensive, often requiring multiple models and significant memory during training. We propose a confidence-based self-distillation approach that outperforms state-of-the-art models by utilizing only previous iteration data storage during training, without requiring extra computation or memory usage during testing. Our approach calculates the loss between the previous and current iterations within a batch using a dynamic confidence coefficient. To evaluate the effectiveness of our approach, we conduct comprehensive experiments on the task of polyp segmentation. Our approach outperforms state-of-the-art models and generalizes well across datasets collected from multiple clinical centers. The code will be released to the public once the paper is accepted.
 
\end{abstract}

\maketitle

\section{Introduction}
\begin{figure*}
	\centering
	\includegraphics[width=0.99\textwidth, height=0.3\textheight]{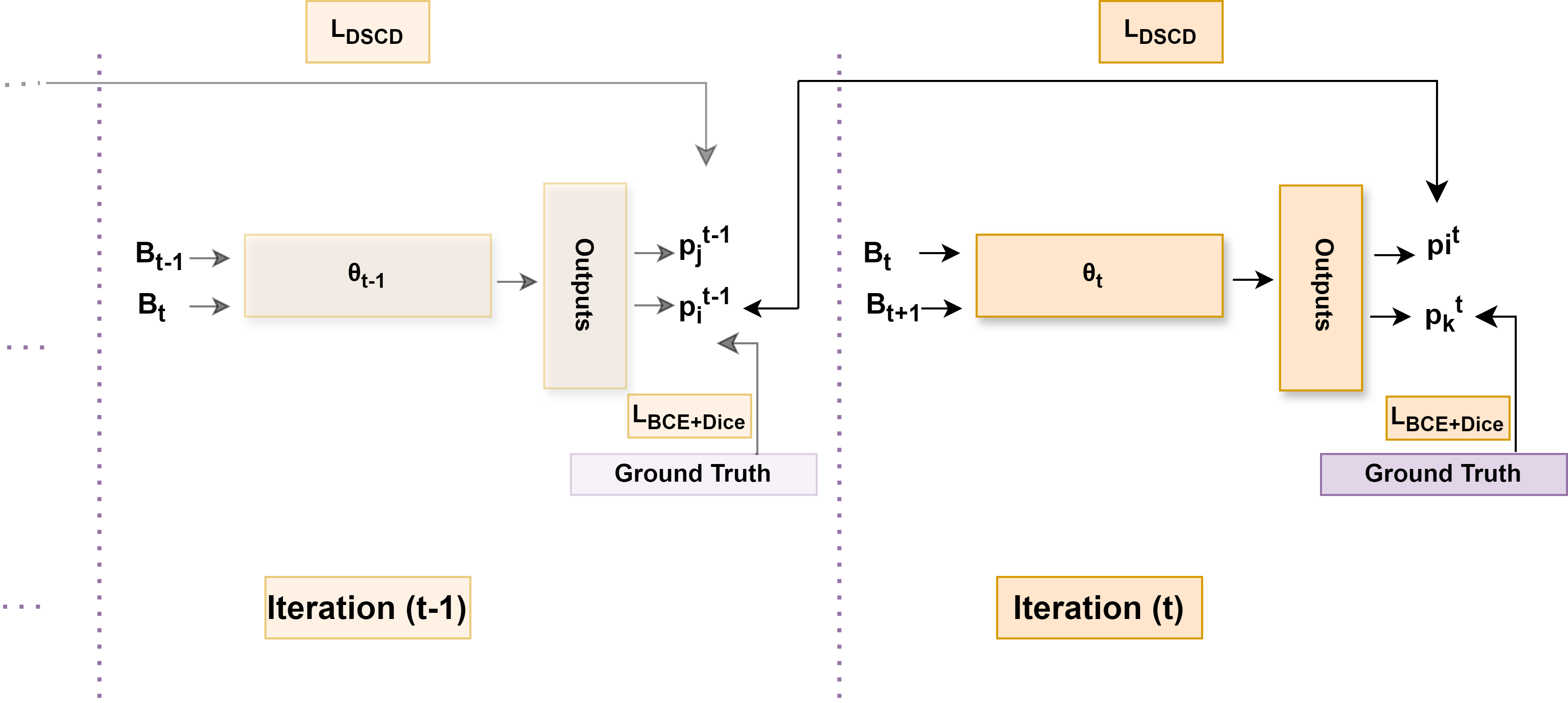}
	\caption{The figure demonstrates our proposed DCSD (Dynamic Confidence-Based Self-Distillation) approach in detail. DCSD calculates the loss between the previous and the current iterations within a batch using a dynamic confidence coefficient. $B_t$, $\theta_t$, and $p^t$ represent the batch, model weights, and prediction at the $t$-th iteration, respectively.}
	\label{dcsdoverview}
\end{figure*}

Colorectal cancer ranks as the third most commonly diagnosed and the second deadliest form of cancer worldwide, according to the World Health Organization (WHO) \cite{who}. Polyps, abnormal tissue growths along the colon lining, can develop into malignant tumors if not detected and removed in time. Despite advances in medical imaging, studies show that between 14\% and 30\% of polyps may go undetected during colonoscopy, depending on their type and size \cite{kvasir}. For this reason, identifying polyps in their early stages is essential to reduce the risk of their progression into colorectal cancer. Deep learning models are increasingly applied to the problem of automatic polyp segmentation, improving detection accuracy and efficiency while reducing the risk of missed polyps. While state-of-the-art models demonstrate strong performance, their reliance on a high number of parameters can result in overfitting, making it difficult for them to generalize across diverse datasets, especially when the data varies in terms of population or imaging conditions. Knowledge distillation approaches address this problem by transferring knowledge from a large, complex model (the teacher) to a smaller, more efficient model (the student).  Due to their reduced complexity, smaller models are less prone to overfitting and generally demonstrate better generalization across diverse datasets while retaining much of the accuracy of the teacher. As a result, a smaller model can be deployed during testing, achieving high performance while requiring fewer resources. Despite this advantage, these methods require training both a teacher and a student model, which can be time-consuming and computationally demanding. As a solution, self-distillation approaches have been proposed. Self distillation methods train a single model by using its own past predictions as a form of guidance, effectively learning from previous epochs or iterations without relying on an external teacher. While self-distillation reduces the need for multiple models, it may still lead to high memory usage, as it often involves storing intermediate outputs (soft targets) for each training instance. Moreover, as the model's earlier predictions may no longer be accurate or relevant due to changes in the data distribution or the underlying patterns in the data over time, the learning process can reinforce these inaccuracies in subsequent training iterations. In this work, we propose a confidence-based self-distillation approach that outperforms state-of-the-art models and retains outputs from the previous iteration only during training, without requiring extra computation or memory usage during testing. Our proposed approach DCSD (Dynamic Confidence-Based Self-Distillation) calculates the loss between the previous and the current iterations within a batch using a dynamic confidence coefficient. This approach improves the model's reliability, consistency, and ability to generalize effectively across diverse datasets. Details of our approach are shown in Figure \ref{dcsdoverview}. 

\begin{figure*}
	\centering
	\includegraphics[width=0.99\textwidth]{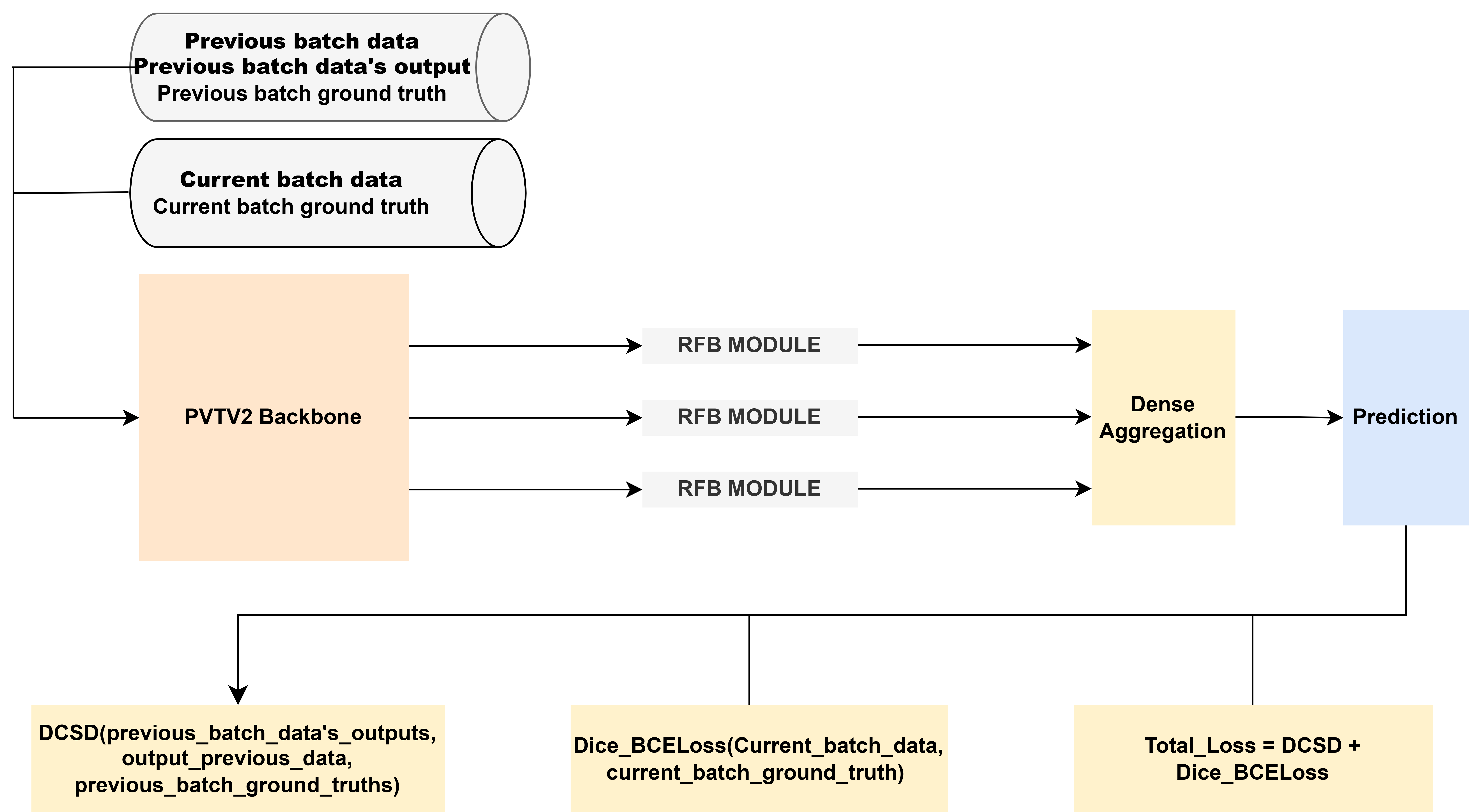}
	\caption{The overview of our architecture and the novel DCSD approach. PVTV2 represents  "Pyramid Vision Transformer" backbone which reduces computational cost through progressively smaller feature map sizes while enhancing generalization across different image sizes and resolutions via multi-scale feature extraction. RFB modules represent Receptive Field Block. Dense aggregation represents deep layer aggregation which aggregates features from multiple layers. In the bottom left corner, DCSD approach which calculates the loss between the previous and current iterations within a batch using a dynamic confidence coefficient is shown.}
	\label{modeloverview}
\end{figure*}

Additionally, in order to further test our approach for the problem of polyp segmentation, we developed a new architecture incorporating a robust backbone and well-established state-of-the-art modules. We use the Pyramid Vision Transformer (PVT) architecture \cite{pvt} as our backbone. While transformer architectures are generally computationally demanding, the Pyramid Vision Transformer (PVT) mitigates this by progressively reducing the spatial resolution of feature maps. It also improves generalization across varying image scales and resolutions through multi-scale feature extraction. We extract three layers from the backbone and feed them into the Receptive Field Block (RFB), which captures diverse spatial patterns by combining features from multiple receptive fields, improving both the robustness and discriminative capacity of the extracted features. Following this, we replace skip connections with layer aggregation, which further improves model performance by integrating features across layers, allowing the model to concurrently exploit fine-grained details and semantic abstractions, ultimately improving accuracy. The overview of our architecture with DCSD is shown in Figure \ref{modeloverview}.

We conducted comprehensive experiments to evaluate our model: First, we trained our model on five different datasets and assessed its performance using a separate, independent dataset. The five datasets we used for training were collected from Ambroise Paré Hospital (Paris), Istituto Oncologico Veneto (Padova), Centro Riferimento Oncologico (IRCCS), Oslo University Hospital (Oslo), and John the Radclife Hospitals (Oxford) \cite{datac6}. We tested our model on the dataset collected from the University of Alexandria, (Alexandria, Egypt). We compared our model with an extensive benchmark which consists of state-of-the-art segmentation models. Secondly, we conducted an ablation study to show the effectiveness of our proposed DCSD approach: We trained the state-of-the-art polyp segmentation specific models TransNetR \cite{transnetr} and ShallowNet \cite{shallownet} as well as our proposed model using a base model, conventional self-distillation, and finally our proposed DCSD approach. We compared the performance of these models using Dice, IoU (Intersection over Union) metrics as well as Precision and Recall. The DCSD approach consistently outperformed both base and self-distillation models on the data\_c6 dataset on Dice and IoU metrics. Then we conducted a second ablation study across various datasets: We trained models with and without our proposed DCSD approach on Kvasir-SEG and CVC-ClinicDB and tested them on Kvasir \cite{kvasir}, CVC-ClinicDB \cite{cvcclinic}, EndoScene \cite{endoscene}, ETIS \cite{etis}, BKAI-IGH \cite{bkai}, and CVC-ColonDB \cite{cvccolon} datasets. Our DCSD approach achieved superior results on the EndoScene, ETIS, and BKAI-IGH datasets. Finally, we further demonstrated the superiority of soft confidence compared to hard confidence through our third ablation study.

Our paper is organized as follows: In section 2, we review the literature on state-of-the-art medical image segmentation models with a focus on polyp segmentation. In addition to this, we review knowledge distillation and self distillation methods. In section 3, we present our proposed DCSD approach and provide a brief overview of the architecture we employed. In section 4, we provide information about the experiments and the datasets, along with the metrics used in these experiments. In section 5, we share our results and compare our model’s performance to the state-of-the-art models. In section 6, we provide a brief conclusion of our work and discuss our findings.

\section{Related Works}

This section reviews recent advances in medical image segmentation and knowledge distillation, highlights the limitations of state-of-the-art models, and explains how our proposed approach addresses these limitations.

\subsection{Image Segmentation}

U-Net, introduced by Ronneberger et al.~\cite{unet}, has become a foundational architecture in medical image segmentation due to its symmetric encoder-decoder design with skip connections that preserve spatial information and enable multiscale feature integration. Its success inspired several variants such as Attention U-Net introduced by Oktay et al.~\cite{attentionunet}, which incorporates attention mechanisms to focus on relevant features, UNet++ by Zhou et al.~\cite{unet++} and UNet3 by Huang et al.~\cite{unet3p}, which improve feature fusion by redesigning skip connections. UNet++ introduces nested dense paths, while UNet3+ connects all encoder and decoder stages to enhance multi-scale representation. Despite advances, many models struggle with maintaining both accuracy and efficiency. SegNet, introduced by Badrinarayanan et al.~\cite{segnet}, addressed the efficiency problem by using pooling indices for upsampling, reducing memory usage.FCN, proposed by Long et al.~\cite{fcn}, mitigated spatial resolution loss by using convolutional layers instead of fully connected ones and incorporating skip connections to refine predictions.PSPNet, introduced by Zhao et al.~\cite{pspnet}, and DeepLabV3+, introduced by Chen et al.~\cite{deeplabv3+}, improved segmentation boundaries by capturing multi-scale context, though at the cost of increased inference time. Hybrid models like ResUNet, introduced by Zhang et al.~\cite{resunet}, improved semantic representation and localization by integrating ResNet with U-Net. Specifically targeting polyp segmentation, SANet, introduced by Wei et al.~\cite{shallownet}, leverages a lightweight design that incorporates shallow attention mechanisms to focus on important features while utilizing a color exchange module to improve the detection of small polyps. Additionally, SANet addresses the issue of class imbalance by implementing a probability correction strategy, which ensures more accurate segmentation, while maintaining real-time performance suitable for colonoscopy. Similarly, TransNetR, introduced by Jha et al. ~\cite{transnetr} combines transformer-based representations with a ResNet50 backbone and a three-decoder setup, offering an efficient approach for polyp segmentation. While SANet focuses on lightweight design and real-time performance through shallow attention mechanisms, TransNetR introduces a more complex architecture aimed at capturing long-range dependencies through transformers. Building on the notion of efficiency, Wu et al.~\cite{pd} proposed a cascaded partial decoder based on the observation that early encoder layers contribute redundant features. Their model excluded low-level layers from attention modules, improving performance without increasing computational load. Extending this idea, Fan et al.~\cite{pranet} introduced Pranet, which utilizes a partial decoder strategy to refine segmentation boundaries. By incorporating attention mechanisms and multi-scale feature aggregation, Pranet achieved state-of-the-art accuracy in polyp segmentation, maintaining efficient performance even in real-time applications. Zhao et al. \cite{msnet} introduced a novel deep learning architecture called Multi-Scale Subtraction Network (MSNet) for automatic polyp segmentation in colonoscopy images. MSNet effectively captures multi-scale difference information using Subtraction Units, reducing feature redundancy typical in U-Net-based models and enabling more accurate and boundary-refined segmentation results. Huang et al. \cite{hardnet} presented HarDNet-MSEG, a lightweight encoder-decoder architecture that integrates the efficient HarDNet68 \cite{hardnet_backbone} backbone with a cascaded partial decoder and reverse attention modules to enable fast and precise polyp segmentation. These developments highlight a key challenge in medical image segmentation: balancing accuracy with computational efficiency. While deep and complex models offer higher accuracy, they are often impractical for real-time or resource-constrained environments, necessitating the exploration of lightweight yet effective alternatives.

\subsection{Knowledge Distillation}

To bridge the gap between performance and efficiency, knowledge distillation (KD) has emerged as a powerful training strategy. Hinton et al.~\cite{kd} introduced the concept of transferring knowledge from a larger teacher model to a smaller student model, enabling the student to mimic the teacher’s behavior and reduce the performance gap. Subsequent work by Romero et al.~\cite{fitnet} enhanced this approach by focusing on intermediate feature representations rather than final outputs, leading to more effective training. Meanwhile, attention-based KD methods, such as the one by Sergey et al.~\cite{attention_kd}, encouraged the student model to replicate the teacher’s attention maps, improving generalization beyond simple soft label imitation. Another variation, relational KD (RKD), proposed by Park et al.~\cite{rkd}, focused on mimicking the relationships between features in the embedding space, rather than individual activations. This approach improved structural understanding and proved more robust across different tasks. Despite these advances, traditional KD methods require training two models simultaneously, which can be computationally expensive. To mitigate this, self-distillation approaches have been proposed. Zhang et al.~\cite{beyourown} introduced the ``Be Your Own Teacher'' framework, where a single model learns from its own predictions across epochs. Furlanello et al.~\cite{bornagain} proposed Born-Again Networks, which iteratively train new models using their predecessors as teachers, maintaining the same architecture. Shen et al.~\cite{sd} offered a more efficient alternative by introducing self-distillation from the last mini-batch, which avoids the need to retain large datasets in memory or train auxiliary models. However, this method can mislead the model if the previous iteration's outputs are noisy or overconfident. To address these shortcomings, we propose a novel Dynamic Confidence-based Self-Distillation approach (DCSD). Unlike traditional KD, DCSD does not require a separate teacher model, and unlike previous self-distillation methods, it introduces a confidence-weighted mechanism when comparing predictions between iterations. By storing only the previous mini-batch and weighing the distillation loss based on prediction confidence, our approach stabilizes training and enhances generalization. Experiments across multiple datasets—including \textit{Data\_C6}, \textit{EndoScene}, \textit{BKAI-IGH}, and \textit{ETIS}—demonstrate that DCSD outperforms both baseline models and previous self-distillation methods, particularly in resource-constrained settings.

\begin{algorithm}
\caption{Dynamic Confident Self Distillation Algorithm}
\label{algoritma-1}
\begin{algorithmic}[1]
    \STATE \textbf{Input:} Training data $imgs$,  $labels$, model.
    \STATE Initialize model, loss function, optimizer
    \STATE \textbf{Output:} Trained model
    \FOR {i, data \textbf{in} enumerate(train\_loader)}
        \STATE imgs, label $\gets$ data
        \STATE out $\gets$ model(imgs)
        \IF {pre\_data \textbf{is not None}}
            \STATE pre\_images, pre\_label $\gets$ pre\_data
            \STATE out\_pre $\gets$ model(pre\_images)
            \STATE dice\_loss $\gets$ dice(out, label)
            \STATE bce\_loss $\gets$ bce(out, label)
            \STATE dcsd\_loss $\gets$ dcsd(out\_pre,pre\_out, pre\_label)
            \STATE total\_loss $\gets$ dice\_loss + bce\_loss +   \( t^2 \) * dcsd\_loss
        \ELSE
            \STATE dice\_loss $\gets$ dice(out, label)
            \STATE bce\_loss $\gets$ bce(out, label)
            \STATE total\_loss $\gets$ dice\_loss + bce\_loss
        \ENDIF
        \STATE pre\_data $\gets$ data
        \STATE pre\_out $\gets$ out

        \STATE optimizer.zero\_grad()
        \STATE total\_loss.backward()
        \STATE optimizer.step()

    \ENDFOR
    
\end{algorithmic}
\end{algorithm}

\begin{algorithm}
\caption{Dynamic Confidence-Based Self Distillation Loss (DCSD)}
\label{algoritma-2}
\begin{algorithmic}[1]
    \STATE \textbf{Input:} The previous mini-batch’s prediction, softened by the temperature value T, is denoted as $\text{pre\_out}$, while the current iteration’s prediction, also softened by T, is denoted as $\text{out\_pre}$.
    \STATE \textbf{Output:} Loss value.
    \STATE criterion = torch.nn.MSELoss()
    \STATE consistency = criterion(pre\_out, out\_pre)
    \STATE confidence-coefficient = 1 - diceloss(pre\_out, pre\_label)
    \STATE loss = consistency * confidence-coefficient
\end{algorithmic}
\end{algorithm}

\section{Proposed Model}
In this section, we first briefly introduce the architecture details and then explain the novel dynamic consistency-based distillation (DCSD) approach. 
An overview of our model and proposed DCSD approach is demonstrated in Figure \ref{modeloverview}. We primarily adopt a encoder-decoder structure using the three encoder layers of Pyramid Vision Transformer \cite{pvt} as the pretrained backbone of our network.We extract three layers from the backbone and feed them into the Receptive Field Block (RFB) \cite{rfbblock}, which enhances the generation of more discriminative and robust features. The RFB block utilizes multiple parallel convolutions with varying kernel sizes to effectively capture multi-scale features, increasing the receptive field and allowing for better contextual understanding.  After that, instead of using skip connections \cite{unet}, we use layer aggregation \cite{aggregation} for better information fusion across layers.

\subsection{Consistency-Based Distillation}
We propose a novel consistency-based distillation approach that calculates the loss between the previous and current iterations within a batch based on the confidence coefficient. Shen et al.~\cite{sd} propose a self mini-batch distillation approach, which can lead to inconsistency during training by directly calculating the loss between the previous and current iterations within a mini-batch. To address this problem, we develop a dynamic confidence coefficient to determine how much information to distill from the previous iteration. Algorithm~\ref{algoritma-1} provides an overview of our approach, while Algorithm~\ref{algoritma-2} illustrates the confidence-based self-distillation loss. The training process of DCSD is visualized in Figure~\ref{dcsdoverview}.

For clarity, we denote the original batch of data sampled in the \( t \)th iteration as \( B_t = \{(x^t_i, y^t_i)\}_{i=1}^n \), and the network parameters as \( \theta_t \). In this context, we substitute \( p_{i}^t \) in Eq.~\ref{dcsd_loss} with the softened labels \( p_{i}^{t-1} \) generated by the same network at the \( (t-1) \)th iteration, specifically \( f \) parameterized by \( \theta_{t-1} \). Additionally, we calculate the confidence score by evaluating the alignment between the softened \( p_{i}^{t-1} \) and the ground truth \( y \), assigning higher confidence to more accurate predictions.

\begin{equation}
\mathcal{L}_{DCSD} = \frac{1}{n} \sum_{i=1}^{n} \left( \operatorname{Dice}(p_{i}^{t-1}, y^{t-1}) \cdot \operatorname{MSE}(p_{i}^t, p_{i}^{t-1}) \right)
\label{dcsd_loss}
\end{equation}

In this formulation, MSE measures the discrepancy between the current prediction \( p_i^t \) and the previous softened prediction \( p_i^{t-1} \), capturing the consistency across iterations. On the other hand, the Dice evaluates the overlap between the previous prediction \( p_i^{t-1} \) and the ground truth \( y^{t-1} \), serving as a confidence score that weights the MSE loss. A higher Dice score implies that the previous prediction was more reliable, and thus should have more influence during distillation. This consistency-based distillation facilitates trustworthy and generalizable outputs by encouraging the model to reinforce only confident past knowledge. Ablation studies demonstrate that our confidence-based approach achieves superior performance on unseen datasets.

\subsection{Theoretical Analysis}

In statistical learning theory, the generalization error \( R(h) \) of a model \( h \) is bounded by:

\[
R(h) \leq \hat{R}(h) + \mathcal{O}\left(\frac{\text{Complexity}(\mathcal{H})}{\sqrt{n}}\right)
\]

where:
\begin{itemize}
    \item \( \hat{R}(h) \) is the training loss,
    \item \( \mathcal{H} \) is the hypothesis space,
    \item \( n \) is the number of training samples,
    \item Complexity($\mathcal{H}$): The ability of a model to adapt to complex relationships in the data.
\end{itemize}

In self-distillation, the model \( h_{\text{SD}} \) learns to predict the same output distribution as its own teacher (the model itself from a previous iteration). This process introduces a regularization effect, reducing the hypothesis space compared to the base model \( \mathcal{H}_{\text{base}} \). The hypothesis space of self-distillation \( \mathcal{H}_{\text{SD}} \) becomes smaller, leading to a tighter generalization bound:

\[
R(h_{\text{SD}}) \leq \hat{R}(h_{\text{SD}}) + \mathcal{O}\left(\frac{\text{Complexity}(\mathcal{H}_{\text{SD}})}{\sqrt{n}}\right)
\]

However, the DCSD approach introduces an additional level of consistency across iterations based on prediction confidence. This means that DCSD not only forces the model to be consistent with its own predictions, but also encourages the model to focus more on regions where its predictions are most confident. This \textit{confidence-based regularization} significantly reduces the hypothesis space compared to self-distillation, making the model more selective and robust in its predictions:

\[
\text{Complexity}(\mathcal{H}_{\text{DCSD}}) < \text{Complexity}(\mathcal{H}_{\text{SD}})
\]

As a result, DCSD achieves a much tighter generalization bound, indicating improved performance on unseen data:

\[
R(h_{\text{DCSD}}) \leq \hat{R}(h_{\text{DCSD}}) + \mathcal{O}\left(\frac{\text{Complexity}(\mathcal{H}_{\text{DCSD}})}{\sqrt{n}}\right)
\]

By incorporating confidence-based consistency, DCSD operates in a more reliable region of the hypothesis space, reducing uncertainty and overfitting. This leads to better performance on unseen datasets, with more stable and robust predictions compared to self-distillation, without requiring additional computation during inference.

\section{Experiments}
We extensively trained and evaluated our approach for polyp segmentation in colonoscopy and wireless endoscopy images across twelve different public datasets. We also conducted ablation studies to demonstrate the effectiveness of confidence-based self-distillation compared to base models and self-distillation \cite{sd} methods. The datasets we used to evaluate our model and dataset properties are summarized in Table \ref{datasets-info}.

\begin{table}[width=.99\linewidth,cols=4,pos=h]
\caption{Table shows the datasets we used to evaluate our model on several medical image segmentation tasks. ``\# images", ``Image Size" and ``Application" represent how many images there are in the corresponding dataset, the width, and height information of the images, and the applications, respectively.}\label{datasets-info}
\begin{tabular*}{\tblwidth}{@{} LLLL@{} }
\toprule
Dataset & \#images & Image Size & Application\\
\midrule
data\_c1 & 256 & Variable & Colonoscopy \\
data\_c2 & 301 & Variable & Colonoscopy \\
data\_c3 & 457 & Variable & Colonoscopy \\
data\_c4 & 227 & 1920x1080 & Colonoscopy \\
data\_c5 & 208 & Variable & Colonoscopy \\
data\_c6 & 88 & Variable & Colonoscopy \\
Kvasir SEG  & 1000 & Variable & Colonoscopy \\ 
CVC-ClinicDB & 612  & 384x288  & Colonoscopy   \\ 
CVC-ColonDB & 380  & 574x500  & Colonoscopy  \\ 
EndoScene   & 60  & 574x500 & Colonoscopy  \\ 
ETIS  & 196  & 1225x966  & W.Endoscopy \\ 
BKAI-IGH & 1000 & Variable & Colonoscopy \\
\bottomrule
\end{tabular*}
\end{table}

\subsection{Experimental Details}
We followed the experimental setup proposed by Ali et al. \cite{datac6}, using the data\_c1 to data\_c5 datasets for training and data\_c6 for testing. In the second experiment, we followed Fan et al. \cite{pranet} and trained model on Kvasir-SEG and CVC-ClinicDB datasets and tested on Kvasir-SEG, CVC-ClinicDB, CVC-ColonDB, ETIS, EndoScene and BKAI-IGH datasets to show DCSD approach's generalizability across different datasets. We resized all images to 256 × 256 × 3. We trained our model on all datasets for 30 epochs. We set the initial learning rate to 1e-4 and used the AdamW optimizer \cite{adamw}. We used Dice, Binary Cross Entropy and Mean Squared Error loss in all experiments.

\subsection{Evaluation Metrics}

In order to evaluate the performance of our models, we utilized the following metrics: Dice coefficient, Intersection over Union (IoU), Precision, and Recall. These metrics are commonly used for segmentation tasks and provide a comprehensive understanding of the model's accuracy.

\begin{table*}[]
\caption{We compared DCSD model with state-of-the-art methods, including FCN \cite{fcn}, U-Net \cite{unet}, PSPNet \cite{pspnet}, ResNetUNet (ResNet34) \cite{resunet}, DeepLabV3+ (ResNet50 \cite{resnet}) \cite{deeplabv3+}, PraNet \cite{pranet}, ShallowNet \cite{shallownet}, TransNetR \cite{transnetr}, HarDNet-MSEG \cite{hardnet} and MSNet \cite{msnet}  in data\_c6 dataset.}\label{benchmark}
\begin{tabular*}{\linewidth}{@{\extracolsep{\fill}}lcccccccl@{}}
\toprule
Methods & Dice & IoU & Precision & Recall  \\
\midrule
FCN & 0.76 & 0.68 & 0.90 & 0.74  \\
U-Net & 0.63 & 0.55 & 0.76 & 0.66  \\
TransNetR & 0.72 & 0.66 & \textbf{0.93} & 0.70 \\
ShallowNet & 0.76 & 0.70 & \textbf{0.93} & 0.77 \\
HarDNet-MSEG & 0.77 & 0.70 & 0.88 & 0.78 \\
Pranet & 0.78 & 0.72 & 0.92 & 0.79 \\
MSNet & 0.79 & 0.72 & 0.91 & 0.80 \\
PSPNet & 0.80 & 0.72 & 0.88 & 0.79  \\
DeepLabV3+(ResNet50) & 0.81 & \textbf{0.75} & 0.92 & 0.79  \\
ResNetUNet(ResNet34) & 0.79 & 0.73 & 0.92 & 0.78  \\
DeepLabV3+(ResNet101) & \textbf{0.82} & \textbf{0.75} & 0.92 & 0.81  \\
ResNetUNet(ResNet101) & 0.80 & 0.74 & \textbf{0.93} & 0.80  \\

Ours & \textbf{0.82} & \textbf{0.75} & 0.91 & \textbf{0.82} \\
\bottomrule
\end{tabular*}
\end{table*}

\begin{table*}[]
\caption{A comparison of our confidence-based self-distillation (DCSD) approach against the base (Base) and self-distillation (SD) methods, as well as the performance of our model using the Dice and IoU metrics, is presented alongside state-of-the-art polyp segmentation models: TransNetR \cite{transnetr}, ShallowNet \cite{shallownet}, and our proposed model. The results demonstrate that our DCSD approach achieved superior scores compared to both the Base and SD methods using the Dice and IoU metrics. Furthermore, our model utilizing the DCSD approach attained the highest scores across all models evaluated, according to Dice and IoU metrics.}\label{benchmark}
\begin{tabular*}{\tblwidth}{@{} LLLLL@{} }
\toprule
Methods &Dice & IoU &Precision &Recall\\
\midrule
TransNetR Base& 0.7189 & 0.6622 & \textbf{0.9348} & 0.6952 \\
TransNetR with SD &0.7413 & 0.6745 & 0.9129 & 0.7310 \\
TransNetR with DCSD & \textbf{0.7538} & \textbf{0.6882} & 0.8786 & \textbf{0.7508} \\
\midrule
ShallowNet Base & 0.7637 & 0.7030 & 0.9253 &0.7684\\
ShallowNet with SD &0.7696 & 0.7036 & 0.9048 & 0.7891 \\
ShallowNet with DCSD & \textbf{0.8202} & \textbf{0.7567} & \textbf{0.9312} & \textbf{0.8123} \\
\midrule

Base & 0.7864 & 0.7182 & 0.8755 & \textbf{0.8196}  \\
SD & 0.7770 & 0.7132 & 0.8679 & 0.8125  \\
DCSD & \textbf{0.8151} & \textbf{0.7542} & \textbf{0.9144} & 0.8184  \\
\bottomrule
\end{tabular*}
\end{table*}

\begin{table*}
\caption{We carried out an experiment to demonstrate the effectiveness of the DCSD approach on the Kvasir \cite{kvasir}, ClinicDB \cite{cvcclinic}, ColonDB \cite{cvccolon}, ETIS \cite{etis}, EndoScene \cite{endoscene} and BKAI-IGH \cite{bkai} datasets. The results showcasing the performance of our approach compared to the Base and self-distillation (SD) methods are presented using the Dice and IoU metrics.}
\centering
\resizebox{\textwidth}{!}{%
\begin{tabular*}{\tblwidth}{@{}l@{\extracolsep{\fill}}ll ll ll ll ll ll@{}}
\toprule
\multirow{2}{*}{Methods}   & \multicolumn{2}{c}{Kvasir} & \multicolumn{2}{c}{ClinicDB} & \multicolumn{2}{c}{ColonDB} & \multicolumn{2}{c}{EndoScene} & \multicolumn{2}{c}{ETIS}  & \multicolumn{2}{c}{BKAI-IGH}\\ 
\cmidrule(lr){2-13}
                           & Dice & IoU & Dice & IoU & Dice & IoU & Dice & IoU & Dice & IoU & Dice & IoU\\ 
\midrule
Base               & 0.8882 & 0.8256 & 0.8804 & 0.8286 & 0.7452 & 0.6631 & 0.8785 & 0.8086 & 0.6863 & 0.6094 & 0.7757 & 0.7009 \\ 
SD           & \textbf{0.9022} & \textbf{0.8418} & \textbf{0.9036} & \textbf{0.8455} & \textbf{0.7681} & \textbf{0.6852} & 0.8680 & 0.7929 & 0.7031 & 0.6300 & 0.7961 & 0.7212 \\ 
DCSD             & 0.8985 & 0.8397 & 0.8994 & 0.8417 & 0.7639 & 0.6794 & \textbf{0.8954} & \textbf{0.8273} & \textbf{0.7121} & \textbf{0.6314} & \textbf{0.8141} & \textbf{0.7386} \\ 
\bottomrule
\end{tabular*}%
}
\label{5_dataset_data}
\end{table*}

\begin{table*}[]
\caption{We conducted an ablation study to investigate whether using soft or hard predictions in comparison with the ground truth yields better confidence estimation in our self-distillation framework. The results show that leveraging soft predictions for this comparison provides a more reliable confidence score, leading to improved distillation performance.}\label{ab2}
\begin{tabular*}{\linewidth}{@{\extracolsep{\fill}}lccccl@{}}
\toprule
Methods & Dice & IoU & Precision & Recall  \\
\midrule
DCSD (Confident T = 1) & 0.7970 & 0.7314 & 0.9039 & 0.8052  \\
DCSD (Confident T = 4) & \textbf{0.8151} & \textbf{0.7542} & \textbf{0.9144} & \textbf{0.8184}  \\

\bottomrule
\end{tabular*}
\end{table*}

\begin{figure*}[t]
    \centering
    \includegraphics[width=\textwidth, trim={0 50 0 50}, clip]{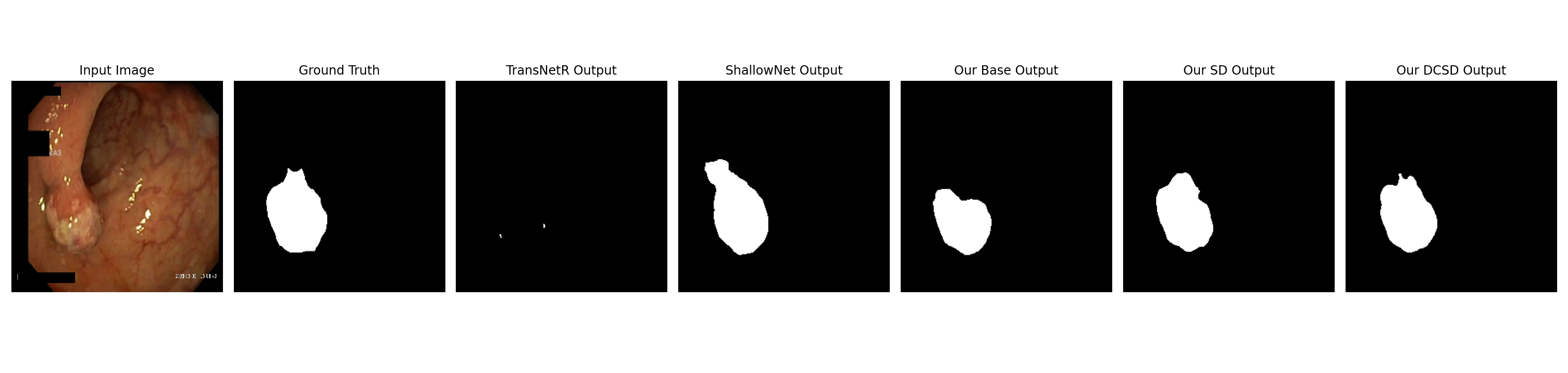}
    \vspace{-5mm}

    \includegraphics[width=\textwidth, trim={0 50 0 50}, clip]{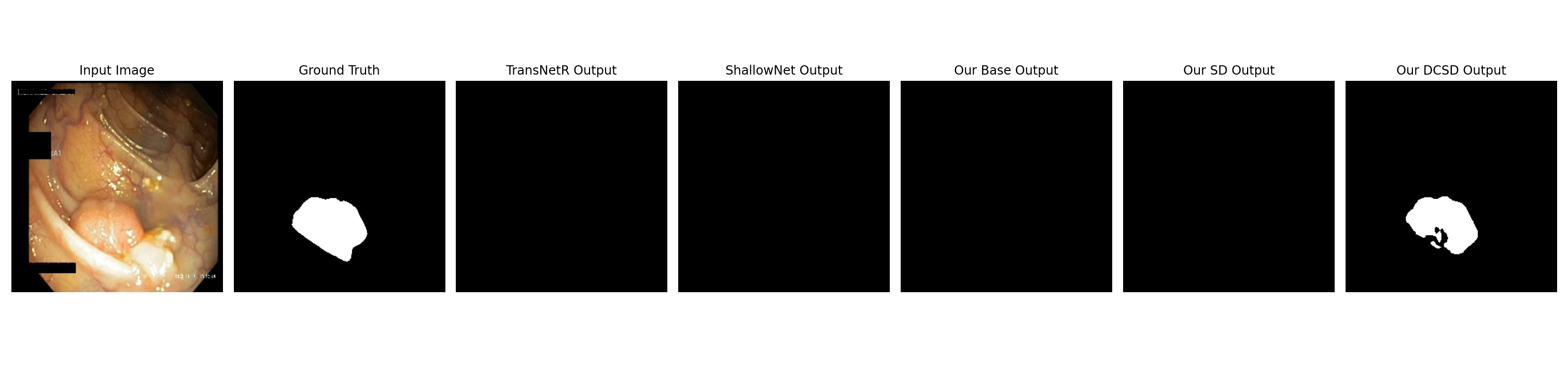}
    \vspace{-5mm}

    \includegraphics[width=\textwidth, trim={0 50 0 50}, clip]{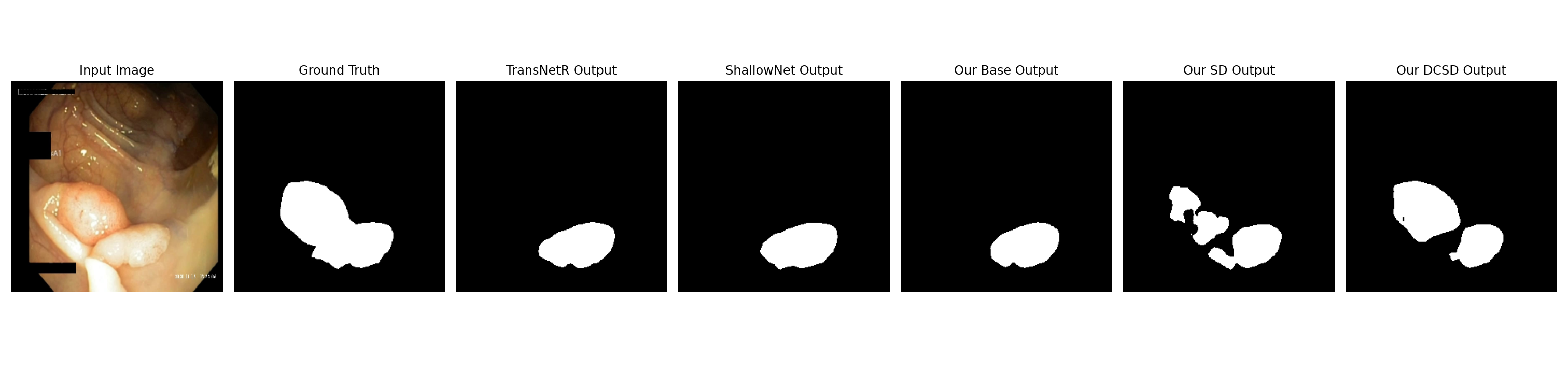}
    \vspace{-5mm}

    \includegraphics[width=\textwidth, trim={0 50 0 50}, clip]{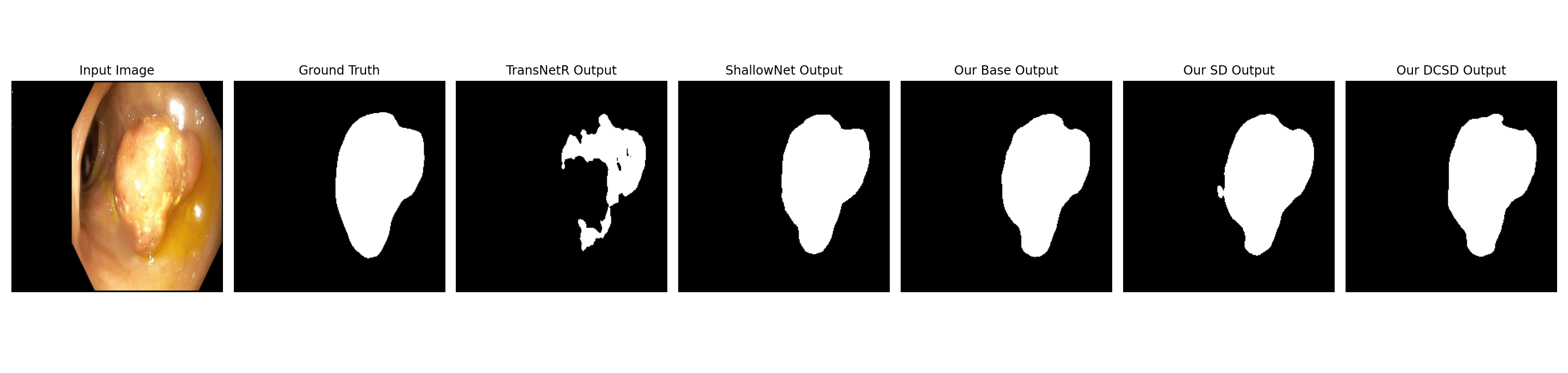}
    \vspace{-5mm}

      \includegraphics[width=\textwidth, trim={0 50 0 50}, clip]{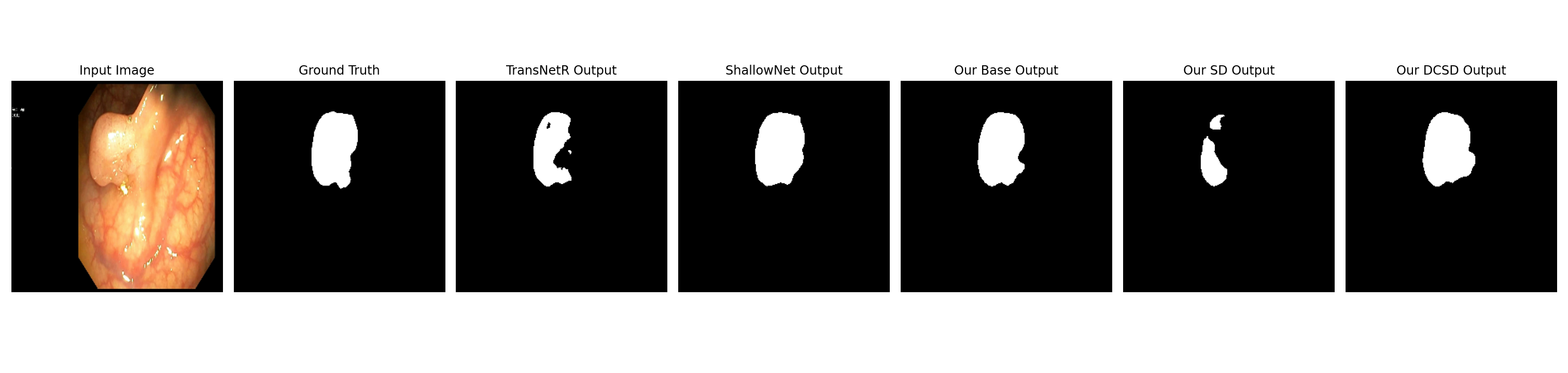}
    \vspace{-5mm}

    \caption{Comparison of model outputs from TransNetR, ShallowNet, our model, our model with SD and our model with the proposed DCSD method. The figure highlights the differences in segmentation performance on the data\_c6 dataset.}
    \label{fig:comparison}
\end{figure*}

\section{Results}
We evaluated our approach on the task of polyp segmentation in colonoscopy and wireless endoscopy images using the data\_c6, Kvasir-SEG, CVC-ClinicDB, CVC-ColonDB, EndoScene,  ETIS and BKAI-IGH datasets. To illustrate the effectiveness of the DCSD approach, we conducted three ablation studies: two of them comparing its performance to that of the base and self-distillation models, and the other examining the differences between soft and hard confidence measures.

\subsection{Polyp Segmentation in Different Models}
We compared DCSD with the benchmark introduced by Ali et al. \cite{datac6}. Table \ref{benchmark} and Figure \ref{fig:comparison} shows our model's results compared to the results of the benchmark models. Our model with DCSD approach outperformed benchmark models on Dice, IoU and Recall metrics.

\subsection{Ablation Study for Polyp Segmentation}
We conducted a second experiment to demonstrate the generalizability and consistency of the DCSD approach across the Kvasir, CVC-ClinicDB, CVC-ColonDB, EndoScene, ETIS and BKAI-IGH datasets. We trained models in Kvasir-SEG and CVC-ClinicDB datasets and tested on Kvasir-SEG, CVC-ClinicDB, CVC-ColonDB, EndoScene, ETIS and BKAI-IGH datasets. In our experiments, the model utilizing the DCSD approach demonstrated superior performance over both the base and self-distillation approaches particularly on unseen datasets. Specifically, the model achieved a \%89.54 Dice score on the EndoScene dataset, \%71.21 Dice score on the ETIS dataset, and a \%81.41 Dice score on the BKAI-IGH dataset. These results underscore the efficacy of the DCSD approach, particularly in the challenging task of polyp segmentation across multiple datasets.  The performance results of the models can be found in Table \ref{5_dataset_data}.

\subsection{Ablation Study for Output Distribution}
Knowledge distillation employs a temperature parameter (T) to the logits. As the temperature value increases, we observe enhanced similarity among the output classes. In this ablation study, we implement the temperature value in the previous batch output to determine the confidence score with the ground truth. We compared this approach to one that does not use the temperature. Specifically, the higher temparature output achieved a \%81.51 Dice score and \%75.42 IoU score on the data\_c6 dataset. These results underscore the efficacy of the  confidence score with higher temperature value approach. The performance results of the models can be found in Table \ref{ab2}.

\section{Discussion and Conclusion}
In this work, we proposed Dynamic Confidence-based Self-Distillation (DCSD), an effective approach to improve model generalization without requiring multiple models or additional computational cost during inference. By leveraging confidence-weighted consistency between successive mini-batches, our approach regularizes training and leads to better segmentation performance, particularly on unseen datasets. Extensive experiments across multiple medical datasets confirm the efficacy of DCSD in outperforming both baseline and previous self-distillation techniques. Furthermore, our ablation studies reveal that using confidence scores derived from soft predictions enhances the reliability of the distillation process. While our approach demonstrates strong generalization and performance, one limitation is its sensitivity to hyperparameters, particularly the temperature used for softening logits and the weight of the distillation loss. Achieving optimal performance requires careful tuning, which may reduce the ease of deployment across different datasets. In future work, we plan to explore adaptive mechanisms for temperature scaling and confidence estimation to enhance the robustness and transferability of DCSD in varying clinical contexts. 

\bibliographystyle{unsrt}
\bibliography{cas-refs}

\end{document}